\documentclass[conference]{IEEEtran}
\IEEEoverridecommandlockouts
\usepackage{ragged2e} 
\usepackage{booktabs,makecell, multirow, tabularx}
\usepackage{cite}
\usepackage{amsmath,amssymb,amsfonts}
\usepackage{algorithmic}
\usepackage{graphicx}
\usepackage{textcomp}
\usepackage{xcolor}
\usepackage[numbers,sort&compress]{natbib}
\def\BibTeX{{\rm B\kern-.05em{\sc i\kern-.025em b}\kern-.08em
    T\kern-.1667em\lower.7ex\hbox{E}\kern-.125emX}}
\begin{document}
\title{LKA-ReID:Vehicle Re-Identification with Large Kernel Attention\\
\thanks{This work was supported in part by the National Natural Science Foundation of China under Grant 62271160 and 62176068, in part by the Natural Science Foundation of Heilongjiang Province of China under Grant LH2021F011, in part by the Fundamental Research Funds for the Central Universities of China under Grant 3072024LJ0803, in part by the Natural Science Foundation of Guangdong Province of China under Grant 2022A1515011527.}
}

\author{\IEEEauthorblockN{1\textsuperscript{st}Xuezhi Xiang}
\IEEEauthorblockA{\textit{Harbin Engineering University} \\
Harbin, China \\
xiangxuezhi@hrbeu.edu.cn}
\and

\IEEEauthorblockN{2\textsuperscript{nd} Zhushan Ma}
\IEEEauthorblockA{\textit{Harbin Engineering University} \\
Harbin, China \\
mazhushan@hrbeu.edu.cn}
\and

\IEEEauthorblockN{3\textsuperscript{rd} Lei Zhang}
\IEEEauthorblockA{\textit{Guangdong University of Petrochemical Technology} \\
	Maoming, China \\
	zhanglei@gdupt.edu.cn}
\and

\and
\IEEEauthorblockN{4\textsuperscript{th} Denis Ombati}
\IEEEauthorblockA{\textit{Harbin Engineering University} \\
Harbin, China \\
deniso2009@gmail.com}

\and

\IEEEauthorblockN{5\textsuperscript{th} Himaloy Himu}
\IEEEauthorblockA{\textit{Harbin Engineering University} \\
	Harbin, China \\
	himaloy@hrbeu.edu.cn}

\and

\IEEEauthorblockN{6\textsuperscript{th} Xiantong Zhen}
\IEEEauthorblockA{\textit{Guangdong University of Petrochemical Technology} \\
Maoming, China \\
zhenxt@gmail.com}

}

\maketitle

\begin{abstract}
		With the rapid development of intelligent transportation systems and the popularity of smart city infrastructure, Vehicle Re-ID technology has become an important research field. The vehicle Re-ID task faces an important challenge, which is the high similarity between different vehicles. Existing methods use additional detection or segmentation models to extract differentiated local features. However, these methods either rely on additional annotations or greatly increase the computational cost. Using attention mechanism to capture global and local features is crucial to solve the challenge of high similarity between classes in vehicle Re-ID tasks. In this paper, we propose LKA-ReID with large kernel attention. Specifically, the large kernel attention (LKA) utilizes the advantages of self-attention and also benefits from the advantages of convolution, which can extract the global and local features of the vehicle more comprehensively. We also introduce hybrid channel attention (HCA) combines channel attention with spatial information, so that the model can better focus on channels and feature regions, and ignore background and other disturbing information. Experiments on VeRi-776 dataset demonstrated the effectiveness of LKA-ReID, with mAP reaches 86.65\% and Rank-1 reaches 98.03\%.
\end{abstract}


\section{Introduction}
With the rapid advancement of intelligent transportation systems and smart cities, vehicle Re-Identification (vehicle Re-ID) technology has emerged as a pivotal research focus \citep{Zheng2016PersonRP}. Vehicle Re-ID entails the analysis and comparison of vehicle images captured by various cameras to discern the identity of the same vehicle, thereby facilitating cross-camera vehicle tracking \citep{9150702}. This technology is crucial in practical applications such as traffic monitoring, intelligent parking management, and traffic accident investigation \citep{2017Learning}.

In recent years, the rapid development of deep learning technology, especially convolutional neural network (CNN), has provided a new solution for vehicle Re-ID. For instance, multi-branch network architectures have been widely applied to extract multi-view features of vehicles, thereby enhancing identification accuracy \citep{2019Partition}. Furthermore, the introduction of attention mechanisms has enabled models to better focus on critical vehicle features, thereby improving identification performance \citep{10.1007/978-3-030-01225-0_30}. Metric learning has effectively differentiated similar vehicles by optimizing specific loss functions, thus enhancing the robustness of re-identification \citep{8100166}. However, a significant challenge faced in  vehicle Re-ID tasks is the high similarity between different vehicles.This is primarily due to the fact that vehicles from the same manufacturer often share many similar attributes in appearance, such as color and model. In the real world, there exists a large number of visually similar vehicles, underscoring the importance of addressing inter-class high similarity issues to enhance the performance of vehicle recognition tasks. To effectively identify vehicles, it is necessary not only to capture global discriminative information but also to focus on local features such as inspection stickers and vehicle logos.

\begin{figure*}[ht]
	\centering {\includegraphics[width=\textwidth]{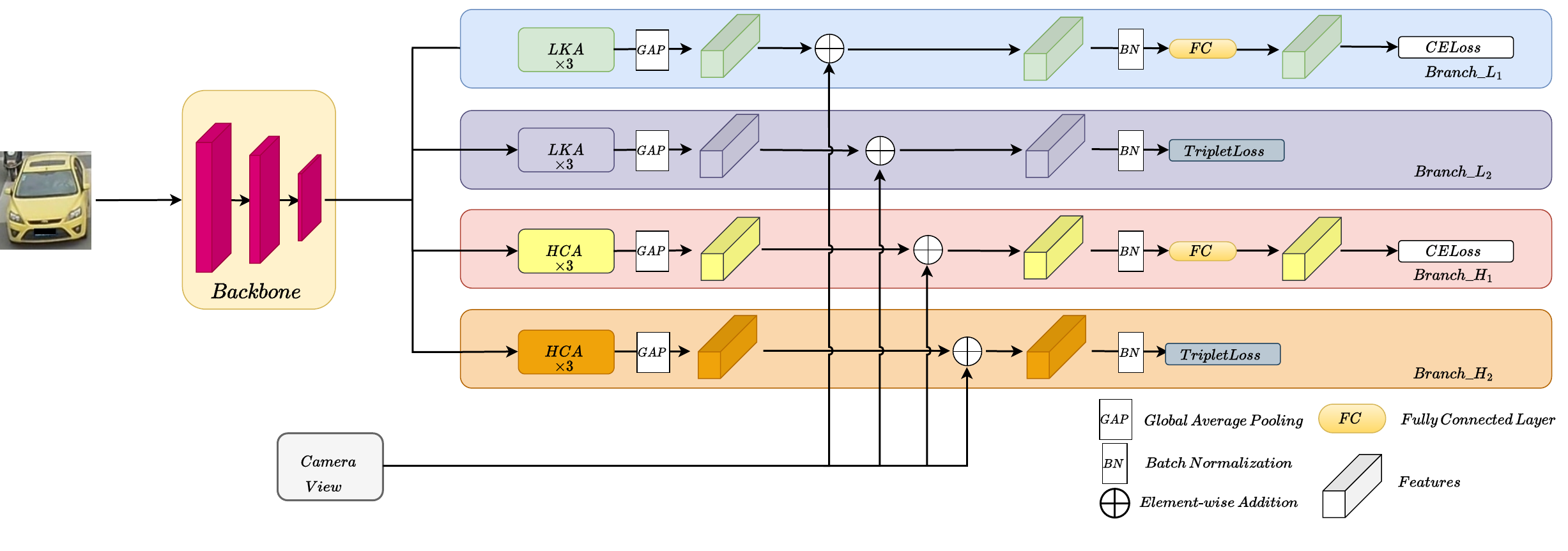}}
	\caption{The overall architecture of our proposed LKA-ReID.}
	\label{fig:my_label1}
\end{figure*}
 
At present, some methods \citep{10422175,Tu2022DFRSTDF,9647974} of vehicle Re-ID extract the global and local information of the vehicle by using convolution and self-attention. However, while the combination of convolution and self-attention mechanisms has significantly improved a vehicle's ability to re-identify, there is still room for further improvement. Convolution is limited by the size of the receptive field. Especially when dealing with long-distance dependencies and global context information, traditional self-attention mechanisms are limited by their computational complexity. For this reason, the recent research trend has turned to the introduction of large kernel attention mechanisms. ConvNeXt \citep{9879745} achieved commendable performance in many visual tasks by leveraging the benefits of large-kernel convolutions. Guo et al.\citep{guo2023visual} proposed a novel linear attention mechanism, dubbed large kernel attention, to facilitate self-adaptive and long-distance correlations in self-attention while circumventing its limitations. Lau et al.\citep{2024Large} proposed a large separable kernel attention that decomposes 2d convolutional kernels of deep convolutional layers into cascaded horizontal and vertical 1d kernels. Inspired by these methods, we propose LKA-ReID with large kernel attention, which introduces LKA to extract the global and local features of the vehicle in the vehicle Re-ID task, so as to effectively solve the challenge caused by view changes.

In addition, inter-vehicle discrimination information and background interference information are the key factors affecting vehicle re-recognition performance. The channel attention mechanism can focus on discriminative information while ignoring intrusive information such as background. For instance, SENet \citep{8578843}, the pioneering channel attention mechanism, was successfully implemented in image classification tasks. Each channel's features are condensed via global average pooling, followed by the generation of channel weights through a fully connected layer, and ultimately adjusted through scaling operations. ECA-Net \citep{9156697} introduced an efficient channel attention mechanism devoid of a fully connected layer, markedly reducing computational complexity. Channel weights are derived by substituting the fully connected layer with a one-dimensional convolution operation, maintaining channel dependencies without a substantial increase in computational load. The Convolutional Block Attention Module (CBAM) \citep{10.1007/978-3-030-01234-2_1} concurrently considers the attention of each feature channel and the feature space, establishing the correlation between each feature channel and the feature space simultaneously. Wan et al.\citep{WAN2023106442} proposed a mixed local channel attention mechanism, capable of amalgamating channel information and spatial information, along with local and global information, to elevate network expressiveness. Inspired by \citep{WAN2023106442}, we introduce hybrid channel attention (HCA) to adaptively weight the spatial and channel dimensions of the feature map, so that the model can focus on the key vehicle's feature regions and channels more effectively, and suppress background and other interference information.

The contributions of this paper are mainly described in the following aspects:

\begin{enumerate} 
	\item We propose LKA-ReID with large kernel attention, which introduces LKA to capture long-distance dependencies, so as to cope with the challenges brought by the change of vehicle appearance and viewpoint.
	\item We introduce HCA that combines channel attention with spatial information, so that the model can better focus on channels and vehicle's key feature regions.
	\item We conducted comprehensive experiments on VeRi-776 dataset. The results show that our method achieves competitive performance, with mAP reached 86.65\% and Rank-1 reached 98.03\%. 
\end{enumerate}
\section{Method}
\label{method}
\subsection{Network Architecture}
The overall architecture of LKA-ReID is shown in Fig. \ref{fig:my_label1}, which is a four-branch network. A single frame of the vehicle image is fed into the LKA-ReID network, and each of the four branches generates 2048 dimensional features. Among them, $Branch\_L_{1}$ and $Branch\_L_{2}$ obtain the global and local features of the vehicle by three LKA modules, which improves the ability to describe the vehicle characteristics. $Branch\_H_{1}$ and $Branch\_H_{2}$ by three HCA modules focus on key features area and channel. $Branch\_L_{1}$ and $Branch\_H_{1}$ use cross-entropy losses to classify samples belonging to different classes, and $Branch\_L_{2}$ and $Branch\_H_{2}$ use triple losses to optimize feature distances between classes. In addition, we configure supplementary metadata in the same way as the baseline \citep{10422175}, including camera ID and vehicle view.

In the inference stage, the output features of all branches are connected in series as the final feature representation. Normalized feature vectors are obtained by L2-norm. Cosine similarity of query and gallery is calculated, and the best matching vehicle is determined according to the similarity score.

\subsection{Large Kernel Attention}
\begin{figure*}[ht]
	
	\centering {\includegraphics[width=\textwidth ]{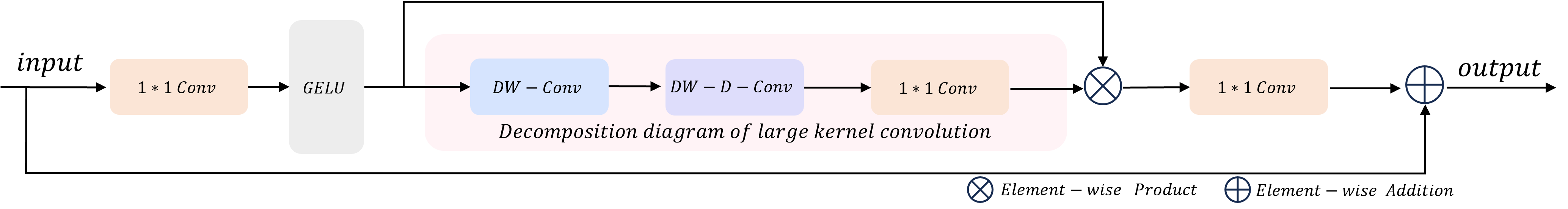}}
	\caption {Large Kernel Attention.}
	\label{fig:my_label2}
\end{figure*}
We introduce LKA \cite{guo2023visual} in the vehicle Re-ID task. As shown in Fig. \ref{fig:my_label2}, a large kernel convolution can be divided into three components: depth-wise convolution, depth-wise dilation convolution and $1\times1$ convolution. Depth-wise convolution captures local spatial information relevant to vehicle features.
Depth-wise dilation convolution captures long-distance spatial dependencies using dilated convolutions, which is particularly useful for identifying vehicles from different viewpoints or distances.
$1\times1$ convolution processes the channels to refine the feature representations, focusing on the most informative vehicle attributes such as color and model. Specifically, the large kernel convolution of $K\times K$ is cleverly replaced a $(2d-1) \times (2d-1)$ depth-wise convolution, a $[\frac{K}{d}] \times [\frac{K}{d}]$ depth-wise dilation convolution with dilation $d$ and a $1 \times 1$  convolution. This decomposition method significantly reduces computational cost and parameters while retaining the ability to capture long-distance dependencies. After extracting long-distance dependencies. As demonstrated in Fig. \ref{fig:my_label2}, LKA can be described  as 	
\begin{equation} \ F^{\prime} =Conv_{1 \times 1}(GELU(F)), \ \end{equation}
\begin{equation} \ Attention =Conv_{1 \times 1}(DW\mbox{-}D\mbox{-}Conv(DW\mbox{-}Conv(F^{\prime}))), \ \end{equation}
\begin{equation} \ Output =Conv_{1 \times 1}( Attention \otimes F^{\prime})+F, \end{equation}
where $F$ is the input vehicle feature. $GELU$ is an activation function. $Attention $ denotes an attention map. The values in the attention map represent the importance of each feature. The large kernel attention combines the advantages of convolution and self-attention, makes full use of the structural information, obtains the global and local features of the image, and captures long-distance dependencies. The computational complexity of self-attention is $O(n^2)$, while the computational complexity of large kernel attention is $O(n)$.
\begin{figure*}[ht]
	
	\centering {\includegraphics[width=\textwidth]{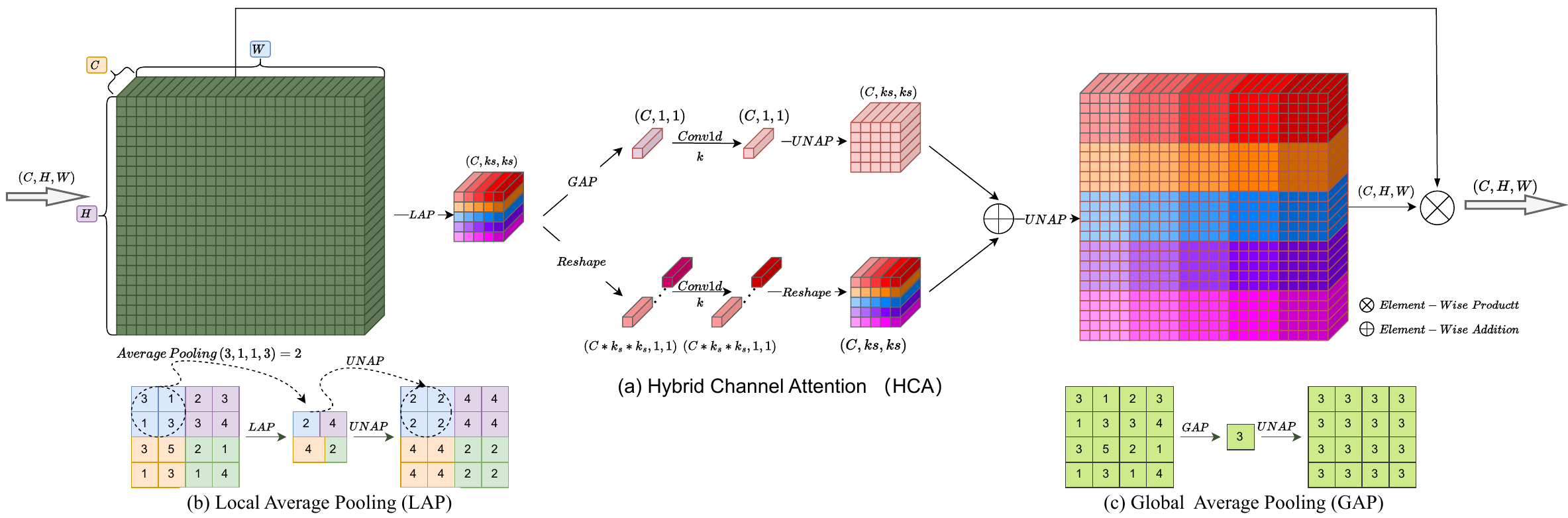}}
	\caption{(a) Hybrid Channel Attention; (b) Local Average pooling; (c) Global Average pooling.}
	\label{fig:my_label4}
\end{figure*}
\subsection{Hybrid Channel Attention}

In order to improve the expression ability of the network for the vehicle features, we introduce HCA \citep{WAN2023106442}, which is shown in Fig. \ref{fig:my_label4}(a). Firstly, the input vehicle's feature map is converted into a $C\times k_{s}\times k_{s}$ vector through local average pooling (LAP), which the LAP is shown in Fig. \ref{fig:my_label4}(b). The first branch then undergoes global average pooling (GAP) to convert the input into a vector of $C\times 1\times 1$,  which the GAP is shown in Fig. \ref{fig:my_label4}(c). The second branch is converted into a $(C\times k_{s}\times k_{s})\times1\times1 $ vector.
The first branch contains global information, and the second branch contains local spatial information. After 1d convolution processing, the output of the two branches is restored to their original resolution by a anti-pooling operation. The two recovered feature vectors are then fused together to synthesize the different features of the vehicle. This process effectively integrates the overall features and detailed features of the vehicle, and improves the recognition accuracy of the vehicle Re-ID task. We set the $k_{s}$ to 5, and the convolution kernel $k$ of $conv1d $ is proportional to channel dimension $C$, indicating that the cross-channel information only considers the relationship between each channel and its k adjacent channels. The selection of $k$ is based on \citep{9156697}, and the formula is as follows
\begin{equation} \ k =\Phi(C)=\left| \frac{\log_{2}{C}}{\gamma} + \frac{b}{\gamma} \right|_{odd}, \ \end{equation}
where $k$ is the size of the $conv1d$ and $C$ is channel dimension. $\gamma$ and $b$ are both hyper-parameters, the default value is 2. $|\cdot |_{odd}$ indicates that $k$ is only odd, and if $k$ is even, then $k$ plus 1.

\section{EXPERIMENTS}
\label{er}
\subsection{Setting}
We conducted experiments on \textbf{VeRi-776} dataset,which contain more than 50,000 images from 776 vehicles taken by 20 cameras over a 24h period, 37,778 images from 576 vehicles for training, and 13,257 images from 200 vehicles for testing. According to the evaluation protocol in \citep{Paszke2017AutomaticDI}, our model was evaluated using the mean average precision (mAP) and the Rank-1 and Rank-5 precision of the cumulative matching characteristic curve (CMC).
Following baseline\citep{10422175}, we perform a PK sampling strategy. On VeRi-776 $P$=6 and $K$=8, resulting in batch sizes of 48. We used metadata from the VeRi-776 dataset, i.e., the camera ID and vehicle view. 

\subsection{Result Statistics}
We compare with recent methods on VeRi-776 dataset, and the results are shown in Tables \ref{tab:table1}.

Our proposed LKA-ReID delivers commendable performance without relying on additional camera ID and vehicle view information. Our method outperforms HCI-Net \citep{SUN2024293} by achieving a 1.75\% higher mAP, a 1.37\% improvement in Rank-1, and a 0.14\% increase in Rank-5 performance. Incorporating camera ID and vehicle view into our network resulted in significant performance boosts. Specifically, our approach led to a 1.10\% increase in mAP, a 0.06\% enhancement in Rank-1, and a 0.06\% rise in Rank-5.
Our proposed method outstrips our baseline MBR-4B \cite{10422175}, by 1.10\% in mAP, 0.29\% in Rank-1, and 0.05\% in Rank-5. These results demonstrate the effectiveness and superiority of our integrated approach in enhancing the accuracy and robustness of vehicle Re-ID task.
\begin{table}[h]
	\caption{Comparison with other methods on VeRi-776 dataset.}
	\centering
	\renewcommand{\arraystretch}{1.2}
	\setlength{\tabcolsep}{10pt}
	\label{tab:table1}
	\resizebox{\linewidth}{!}{%
		\begin{tabular}{c|clc}
				\toprule[1pt]
			\multirow{2}{*}{\textbf{Method}} & \multicolumn{3}{c}{\textbf{VeRi-776}}                                              \\ \cline{2-4} 
			& \multicolumn{1}{c|}{mAP\( \uparrow \)} & \multicolumn{1}{c|}{Rank-1\( \uparrow \)} & \multicolumn{1}{c}{Rank-5\( \uparrow \)} \\ \hline
			\multicolumn{1}{c|}{FastREID \citep{He2020FastReIDAP}}            & \multicolumn{1}{c|}{81.9}    & \multicolumn{1}{c|}{97.0}      & {98.4}       \\ 
			\multicolumn{1}{c|}{TransREID* \citep{He_2021_ICCV}}            & \multicolumn{1}{c|}{82.3}    & \multicolumn{1}{c|}{97.1}      & {-}       \\ 
			\multicolumn{1}{c|}{HRCN \citep{9711371}}            & \multicolumn{1}{c|}{83.1}    & \multicolumn{1}{c|}{97.3}      & {98.9}       \\ 
			\multicolumn{1}{c|}{ANet \citep{QUISPE202184}}            & \multicolumn{1}{c|}{81.2}    & \multicolumn{1}{c|}{96.8}      & {98.4}       \\ 
			\multicolumn{1}{c|}{MUSP \citep{9922319}}            & \multicolumn{1}{c|}{78.8}    & \multicolumn{1}{c|}{95.6}      & {97.9}       \\ 
			\multicolumn{1}{c|}{TANet \citep{Lian2022TransformerBasedAN}}            & \multicolumn{1}{c|}{80.5}    & \multicolumn{1}{c|}{95.4}      & {98.4}       \\ 
			\multicolumn{1}{c|}{SSBVER \cite{10208396}}            & \multicolumn{1}{c|}{82.1}    & \multicolumn{1}{c|}{97.1}      & {98.4}       \\ 
			\multicolumn{1}{c|}{MBR-4B \citep{10422175}}            & \multicolumn{1}{c|}{84.72}    & \multicolumn{1}{c|}{97.68}      & {98.81}       \\
			\multicolumn{1}{c|}{MBR-4B* \citep{10422175}}            & \multicolumn{1}{c|}{85.63}    & \multicolumn{1}{c|}{97.74}      & {99.05}       \\
			\multicolumn{1}{c|}{HCI-Net \citep{SUN2024293}}            & \multicolumn{1}{c|}{83.8}    & \multicolumn{1}{c|}{96.6}      & {98.9}       \\ 
			\hline
			\multicolumn{1}{c|}{Ours }            & \multicolumn{1}{c|}{85.55}    & \multicolumn{1}{c|}{97.97}     & {99.04}    \\
			\multicolumn{1}{c|}{Ours* }            & \multicolumn{1}{c|}{\textbf{86.65}}    & \multicolumn{1}{c|}{\textbf{98.03}}      & {\textbf{99.10}}       \\ 
			
			\bottomrule[1pt]
			\multicolumn{4}{l}{* indicates the use of extra data.}\\
		\end{tabular}
	}
\end{table}
\subsection{Ablation Analysis}
\begin{figure}[ht]
	
	\centering {\includegraphics[width=\columnwidth ]{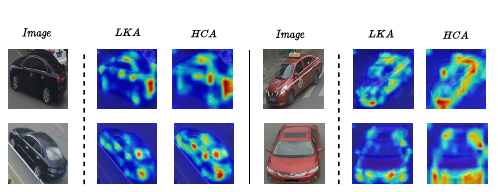}}
	\caption{Visualization of activation maps on VeRi-776 dataset.}
	\label{fig:my_label3}
\end{figure}

\begin{table}[ht]
	\caption{Ablation results for each component of our method on VeRi-776 dataset}
	\centering
	\renewcommand{\arraystretch}{1.2}
	\setlength{\tabcolsep}{4pt}
	\label{tab:table2}
	\resizebox{\columnwidth}{!}{%
		\begin{tabular}{c|cc|ccc|cc}
			\toprule[1pt]
			Method                                     & LKA  & HCA & mAP\( \uparrow \) & Rank-1\( \uparrow \) & Rank-5\( \uparrow \) &  Params(M)   &  GFLOPs  \\ \hline
			\multicolumn{1}{l|}{Baseline}                  &      &     & 84.72   &  97.68     & 98.81&  61.44   &14.44\\
			
			\multicolumn{1}{l|}{}                       &    \checkmark   &     &  85.37   &    97.79   &     98.84      &   61.57  &   13.89        \\
			\multicolumn{1}{c|}{Ours}                    &      &   \checkmark  &   85.30   & 97.74   &  98.86         &  47.29   &   14.25     \\
		
			\multicolumn{1}{l|}{}                        &   \checkmark   &  \checkmark   &  \textbf{85.55}   &   \textbf{97.94}   &     \textbf{99.04}         & \textbf{47.42}     & \textbf{13.7}       \\ \bottomrule[1pt] 
		\end{tabular}
	}
\end{table}
We conducted extensive ablation experiments, as shown in Table \ref{tab:table2}. Firstly, by incorporating the LKA into our baseline, we achieved an improvement of 0.65\% in mAP, with almost no change in the number of parameters, and a reduction of 0.55 GFLOPs. By integrating the HCA into our baseline, we observed an increase of 0.58\% in the mAP. Notably, this improvement came alongside a significant reduction in the number of parameters—approximately 23.03\%. This demonstrates that the HCA not only boosts the performance but also makes the model more parameter-efficient. When both the LKA and the HCA were incorporated into the baseline, we observed significant performance uplifts: a 0.83\% increase in mAP, a 0.29\% rise in Rank-1 accuracy, and a 0.23\% improvement in Rank-5 accuracy. In addition, Fig. \ref{fig:my_label3} shows activation graphs obtained through LKA and HCA using Grad-CAM \citep{8237336}.  Most of the active areas (marked in red) are the lights, windows, front of the car, and the edge of the car.

\section{Conclusion}
\label{con}
In this paper, we propose an effective vehicle Re-ID method named LKA-ReID. The LKA captures long-distance dependencies, enhancing the network's ability to extract comprehensive features. Meanwhile, the HCA efficiently integrates spatial and channel information, allowing the model to concentrate more effectively on critical feature regions and ignore the background interference information. Extensive experimental validation on benchmark datasets demonstrates the superior performance and effectiveness of our proposed method. Although our proposed method improves the performance, the four-branch structure also brings some computational complexity. In the future, we will explore lighter and more efficient network architectures.

\bibliographystyle{IEEEtran}
\bibliography{refs}

\end{document}